\documentclass[11pt,a4paper]{article}
\usepackage[hyperref]{acl2019}
\usepackage{times}
\usepackage{latexsym}
\usepackage{amsfonts}
\usepackage{amsmath}
\usepackage{algorithm}
\usepackage{algpseudocode}
\usepackage{tabularx}
\usepackage{graphicx}

\usepackage{url}

\aclfinalcopy % Uncomment this line for the final submission
\title{Actions Generation from Captions}

\author{Xuan Liang \\
  University of Technology Sydney \\
  \texttt{xuan.liang@student.uts.edu.au} \\\And
  Yida Xu\\
  University of Technology Sydney \\
  \texttt{yida.xu@uts.edu.au} \\}

\date{}

\begin{document}
\maketitle
\begin{abstract}
Sequence transduction models have been widely explored in many natural language processing tasks. However, the target sequence usually consists of discrete tokens which represent word indices in a given vocabulary. We barely see the case where target sequence is composed of continuous vectors,  where each vector is an element of a time series taken successively in a temporal domain. In this work, we introduce a new data set, named Action Generation Data Set (AGDS) which is specifically designed to carry out the task of caption-to-action generation. This data set contains caption-action pairs. The caption is comprised of a sequence of words describing the interactive movement between two people, and the action is a captured sequence of poses representing the movement. This data set is introduced to study the ability of generating continuous sequences through sequence transduction models. We also propose a model to innovatively combine Multi-Head Attention (MHA) and Generative Adversarial Network (GAN) together. In our model, we have one generator to generate actions from captions and three discriminators where each of them is designed to carry out a unique functionality: caption-action consistency discriminator, pose discriminator and pose transition discriminator. This novel design allowed us to achieve plausible generation performance which is demonstrated in the experiments.
\end{abstract}

\begin{figure*} 
  \includegraphics[scale=0.06]{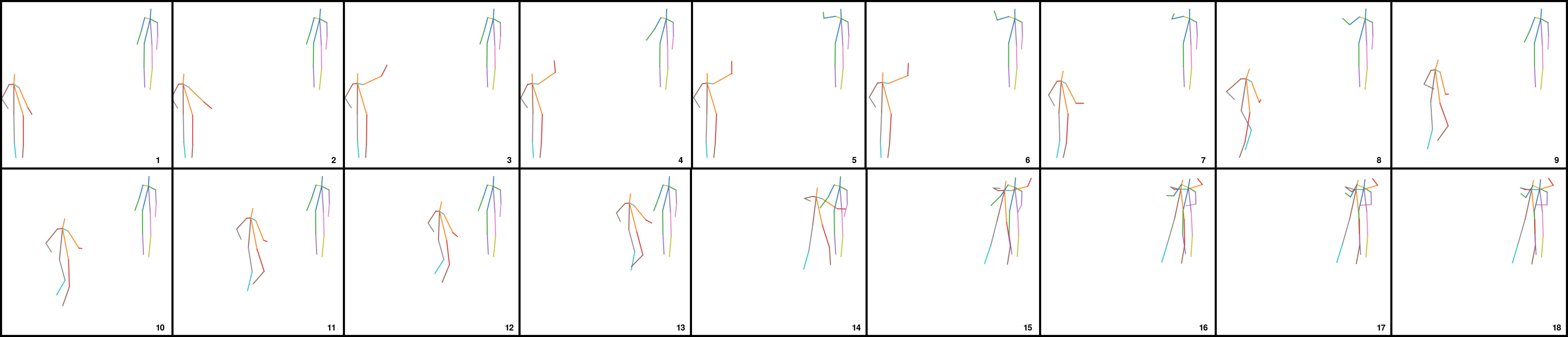}
\caption{One example of Action Generation Dataset. The corresponding caption is: "A waves at B and runs towards B, cause B to wave back. A reaches B and hugs B."}
\end{figure*} 

\section{Introduction}

Sequence transduction models have been extensively studied in many natural language processing tasks, such as neural machine translation \cite{DBLP:journals/corr/SutskeverVL14, DBLP:journals/corr/VaswaniSPUJGKP17} and abstractive text summarisation \cite{DBLP:journals/corr/abs-1803-10357}. Generally, these models have an encode-decoder structure. Within an encoder/decoder, we can flexibly use many deep learning models, such as convolutional neural network (CNN), recurrent neural network (RNN) or the recently proposed Multi-Head Attention (MHA) as their components.

In most public data sets, the target sequence is discrete. A typical example is sentences, which are comprised of vocabulary words to be used as the outputs of a language generation model. Despite the popularity of language applications \cite{DBLP:journals/corr/KumarISBEPOGS15, DBLP:journals/corr/SeeLM17, DBLP:journals/corr/VenugopalanXDRMS14}, there also exist many other settings where the output contains a series of continuous data: for example, video sequence and animation. The ability of how existing sequence transduction models may generate continuous sequence is not well understood. We partly attribute this to the lack of dataset in continuous output. 

To this end, we introduce a new data set: Action Generation Data Set (AGDS) and a task---generating actions from captions. This data set consists of Action-Caption pairs, where each \textit{Caption} is a sequence of words to describe the movement of two people, and its \textit{Action} is comprised of a continuous sequence of body poses depicting its corresponding interactions. Each of the body pose is a dense vector comprising 3D coordinates representing a person's joints. The task is to generate a continuous sequence of \textit{Action} given a discrete sequence \textit{Caption}. Figure 1 shows one example in our data set.

It is impractical to apply the existing discrete sequence generation model directly to our setting due to the four key differences between the two: 

The first key difference is that in discrete output domain, the input and its corresponding output usually has one-to-one relationship. For example, although there exist some small grammatical variations in the outputs of a language model, they should mean the same thing. However, this may not be the case in the continuous case. For example, in our setting, each caption may have multiple reasonable movements. For the word 'walk', two steps forward or three steps to the right are both correct ways to perform the word 'walk'. Instead of a one-to-one data set, we create a one-to-many data set. Traditional sequence transduction model, which has an encoder and a decoder along with attention mechanism, is not sufficient for the introduced task, as we can not assign a unique target sequence to a given input sequence and train it in the usual supervised way. We employ Generative Adversarial Network (GAN) \cite{NIPS2014_5423} to the task to learn the mapping in an adversarial way. 

The second difference is that for discrete output, we usually only need to place an overall discriminator which ensures the global consistency between the input and output sequence, hence we named it \textit{Consistency-Discriminator}. In the continuous output setting, we are generating the \textit{data} instead of an index. Therefore, additional care must be taken to make sure (1) the data at each time-step is making sense, for example, a human pose still obey anatomical rules, (2) the transitions between \textit{data} is also satisfying the constraints of its context. In the case of human motion, they must also obey human motion physics. According to the above argument, we additionally introduce two constraints in our model. One is called the \textit{pose constraint}, which controls how likely a pose will be generated given a caption $p(pose|caption)$. The other is called \textit{pose transition constraint} which controls how likely the next pose will be generated given a caption and a pose at any time step $p(pose_{t}|pose_{t-1}, caption)$. The details of the two constraints will be discussed in section 4.

The third difference is that for continuous sequence, the change between two adjacent time step is relatively small and smooth. This is an inherent characteristic for continuous sequence during its generation. In contrast, indices of two adjacent words in the discrete sequence may vary significantly. We find in our experiments that if we use Multi-Head Attention as our main model component, it leads to over-smoothing problem. We solve it by replacing all layer normalisation \cite{Ba2016LayerN} with batch normalisation \cite{DBLP:journals/corr/IoffeS15} to encourge the difference between adjacent time steps.

The last difference is that in our setting, there is no need to pay special attention to the so-called {\it generator differentiation problem}\cite{DBLP:journals/corr/YuZWY16}, which is solved by gradient policy update \cite{DBLP:journals/corr/YuZWY16}. Because the output sequence is continuous, it can be passed to the discriminator directly.

In summary, in an effort to solve these new challenges introduced by having continuous output generation, we propose a new framework containing the elements of novel solutions which are reflected in our model. Therefore, we organise the rest of the paper as follows: section 2 reviews some of the popular sequence transduction models; section 3 introduces the Action Generation Data Set (AGDS); section 4 describes the proposed general training framework for conditional continuous sequence generation; section 5 demonstrates the experiments and generation performance; section 6 concludes this work. 

\section{Related Works}

Encoder-decoder structure \cite{DBLP:journals/corr/SutskeverVL14, DBLP:journals/corr/ChoMBB14} has become a general framework in sequence transduction models. Encoder usually maps each time step in the input sequence to a new representation, the decoder then generates the output sequence based on these new representations. Attention mechanism \cite{DBLP:journals/corr/BahdanauCB14, DBLP:journals/corr/XuBKCCSZB15} plays an import role in such kind of models, it allows the decoder to focus on different parts of the input sequence when generating the output sequence at each time step. 

Recurrent neural network (RNN), especially long short-term memory \cite{HochSchm97} is the most widely used layer in sequence transduction models. However, the inherent sequential nature of RNN model precludes parallelisation. Google proposed the Transformer \cite{DBLP:journals/corr/VaswaniSPUJGKP17} architecture to eschew recurrence and only based on attention mechanism which significantly improves the training speed. Facebook also proposed their sequence transduction model architecture \cite{DBLP:journals/corr/GehringAGYD17} which based entirely on convolutional neural networks (CNN) to parallelise computation.

Generative adversarial network (GAN) \cite{NIPS2014_5423} has also been employed to language generation task. In \cite{DBLP:journals/corr/YuZWY16,DBLP:journals/corr/WuXZTQLL17,DBLP:journals/corr/YangCWX17}, the generator is the sequence transduction model, and the discriminator is to discriminate between machine generated sequences and the human generated ones.  Policy gradient update and Monte Carlo search are utilised to solve the generator differentiation problem. Moreover, teacher forcing plays a significant role in improving training stability in the adversarial training structure.

Since most of the current popular transactional researches were proposed to solve language modelling problem which is concerning a discrete output, researchers have yet started to systematically explore the continuous output counterpart, which leads to less references we can draw inspirations from. One pioneering method reminiscent to our work is on Conditional video generation \cite{DBLP:journals/corr/abs-1804-08264}. Their goal is to generate video which is comprised of a sequence of images. In here, the paper proposed a model called TGANs-C, which has three discriminators to help generating videos semantically and temporal coherently. The convolutional layer that has proven its success in image-related tasks, is employed as the main component in the proposed model. Although our model was also proposed to solve continuous output generation, however the nature of animation data are in stark contrast to those of video. To begin with, animations are best expressed using human joint positions as oppose to video frames. Therefore, each dimension of a joint-frame at time $t$ corresponds to joint-frame at time $t+1$, which is a property that video frame generation does not enjoy. For this reason, we do not need to add a separate CNN layer to ``extract" feature from the raw input, i.e., the joint position are already ``feature-extracted". Secondly, since we have well-defined features using body joints at each time $t$, we can also take advantage of this and to impose addition constraint such as smoothing constraints between their corresponding dimensions at two consecutive timeframe. People may argue that the same can also be said about between consecutive video frames. However, pixel of the same $x-y$ position across time do not necessary correspond to the same object and placing a smoothing constraint would only blur the image instead of achieving true smoothing. For these reasons, we have proposed a new Discriminator which to incorporate all these natural phenomenons occurring in our animation data set. We also like to emphasis that our novel constraints are not only useful in animation dataset, but they can be trivially applied to other continuous output problems where dimension of each feature correspond to the same thing across time.

\section{Action Generation Data Set}

To study conditional continuous sequence generation task, we introduce a new data set, named Action Generation Data Set (AGDS). This data set contains caption-action pairs. To best cater for our unique goals stated in the introduction and to showcase the one-to-many property of our model, we artificially made our dataset to have each text input corresponding to multiple animation outputs by duplicating the output sequence with the ones with viewing-angle alterations. Of course, further duplicative alterations of the output sequences can be triviality achieved in the future by using frame dropping/addition or by adding noise to joint positions - which we left this open to other researchers experimenting with our dataset. Each \textit{caption} comprised of a sequence of words $\mathbf{x}=(x_{1},...,x_{n})$ describing the interactive movement between two people, where $x_i$ is the index of a given word in the vocabulary. Each \textit{action} $\mathbf{y}=(\mathbf{y}_{1},...,\mathbf{y}_{m})$ has two captured subsequences of poses. One subsequence of poses corresponding to one person, representing the movement. Pose contains the 3D coordinates of 14 joints, which are \textit{head, neck, left shoulder, left elbow, left hand, left thigh, left shin, left foot, right shoulder, right elbow, right hand, right thigh, right shin and right foot}. Then $\mathbf{y}=[\mathbf{y}^{1};\mathbf{y}^{2}]$, where $\mathbf{y}^{i}=(\mathbf{y}_{1}^{i},...,\mathbf{y}_{m}^{i})$ and pose $\mathbf{y}_{j}^{i} \in \mathbb{R}^{3\times14}$. 

All our data are collected by Motion Captions System (OptiTrack), with 120 frames per second. We uniformly sample frames to limit the length of the action sequence. As a given caption may have multiple corresponding actions, we record two caption-action pairs in the data set if two different actions correspond to one same caption.

\begin{table}[t!]
\caption{Statistics of Action Generation Data Set}
\centering
\scalebox{0.95}{
\begin{tabular}{l c l c}
\hline\hline
Attributes & Value & Attributes & Value \\ [0ex]
\hline

Num. of &  & Max Length &  \\ [-1ex]
Unique Captions & \raisebox{1.5ex}{229} & of Captions & \raisebox{1.5ex}{43} \\ [1ex]

Num. of Caption&  & Max Length &  \\ [-1ex]
-Action Pairs & \raisebox{1.5ex}{358} & of Actions & \raisebox{1.5ex}{26} \\ [1ex]

Num. of Pairs&  &  &  \\ [-1ex]
after Rotation & \raisebox{1.5ex}{12888} & \raisebox{1.5ex}{Vocab Size} & \raisebox{1.5ex}{235} \\ [1ex]

\hline
\end{tabular}}
\end{table}

Table1 shows some key statistics of the Action Generation Data Set.  It contains 229 unique captions and 358 caption-action pairs. We also rotate the angle of each action to expand the data set. For each caption-action pair, we randomly sample $36$ angles between $0$ and $2\pi$, then rotate the action by those sampled angles. After rotation, the dataset has 12888 caption-action pairs. Besides, the maximum length of captions is 43 and the maximum length of actions is 26 (3 frames per second). We restrict the caption vocabulary size to 235 as this is a relatively small data set.

\section{Model Architecture}
As introduced in the last section, AGDS is a transductional dataset with one-to-many relationship. Therefore, we employ an adversarial training architecture, the introduction of noise helps generating diverse actions given a caption. The generator $\mathbf{G}$ is chosen to have an encoder-decoder structure. It aims to generate an action $\mathbf{y}=(\mathbf{y}_{1},...,\mathbf{y}_{m})$, where $\mathbf{y}_{i} \in \mathbb{R}^{d_{f}}$, given an input caption $\mathbf{x}=(x_{1},...,x_{n})$. The goal of the discriminator $\mathbf{D}$ is to discriminate model generated actions from human performed actions. 

\subsection{Generator}
As the use of Transformers \cite{DBLP:journals/corr/VaswaniSPUJGKP17}  has become ubiquitous in recent NLP researches \cite{DBLP:journals/corr/abs-1804-09541,DBLP:journals/corr/abs-1810-04805}, we have incorporated this model to our action generation task using our modified network structure. The structure of the generator is shown in Figure 2. To make the paper to be self-contained, first we describe two main components in generator: Multi-Head Attention (MHA) layer and Point-Wise Fully Connected Feed-Forward (PFCF) layer.

\begin{figure} 
  \includegraphics[scale=0.038]{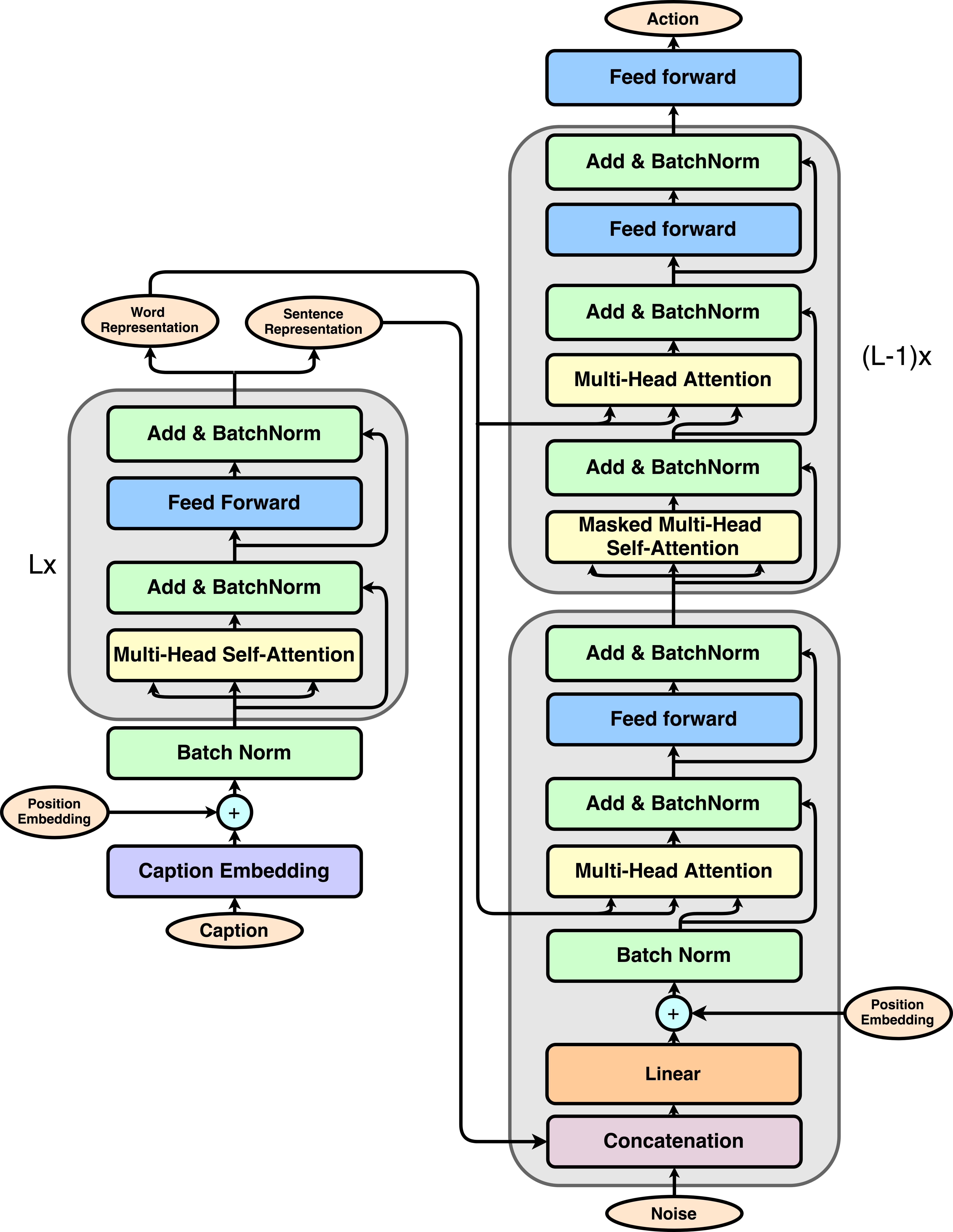}
\caption{Generator Structure}
\end{figure} 

\textit{Multi-Head Attention Layer} has three inputs $Q$, $K$ and $V$. Multi-head attention is concatenated by $H$ scaled dot-product attention.  For each scaled dot-product attention, the three input matrixes are linearly projected to three matrixes $Q_p$, $K_p$ and $V_p$ seperately with dimension $d_{h}$. The attention is the commonly used dot product similarity calculation:

\begin{equation} \label{eq1}
\begin{split}
head &= Attention(Q_p,K_p,V_p) \\
&= softmax(\frac{Q_{p}K_{p}^{T}}{\sqrt{d_{h}}})V_{p}
\end{split}
\end{equation}

\noindent $\sqrt{d_{h}}$ is used as a scaling factor to prevent large values of dot product. We calculate this scaled dot-product attention $H$ times with different weights and concatenate the $H$ outputs together, then project it back to dimension $d$:

\begin{equation} \label{eq2}
\begin{split}
MultiHead&(Q,K,V) = \\
&Concat(head_{1},..., head_{h})W^{o} 
\end{split}
\end{equation}

\begin{equation} \label{eq3}
\begin{split}
head_{i} &= \\
&Attention(Q_{p}W_{i}^{Q_{p}}, K_{p}W_{i}^{K_{p}}, V_{p}W_{i}^{V_{p}} )
\end{split}
\end{equation}

\noindent Projection matrixes $W_{i}^{Q_{p}} \in \mathbb{R}^{d \times d_h}$, $W_{i}^{K_{p} \in \mathbb{R}^{d \times d_h}}$, $W_{i}^{V_{p}} \in \mathbb{R}^{d \times d_h} $ and $W^{o} \in \mathbb{R}^{hd_h \times d} $.

\textit{Position-Wise Fully Connected Feed-Forward Layer}, as its name said, it applies fully connected feed-forward layer to each position separately and identically. It consists of two linear layer with a GELU \cite{DBLP:journals/corr/HendrycksG16} activation function in between:

\begin{equation} \label{eq4}
FFN(x) = GELU(xW_1+b_1)W_2+b_2
\end{equation}

\noindent where $W_1 \in \mathbb{R}^{d*d_p}$ and $W_3 \in \mathbb{R}^{d_p*d}$.

~\\
\noindent \textbf{Encoder} is shown in the left part of Figure 2, the encoder is almost the same as the encoder in Transformers. First, we convert caption $\mathbf{x}=(x_{1},...,x_{n})$ to word embeddings as $\mathbf{e}=(\mathbf{e}_1,...,\mathbf{e}_n)$, where $\mathbf{e}_i \in \mathbb{R}^d$ is a row of embedding matrix $\mathcal{E}\in \mathbb{R}^{V \times d}$ and $V$ denotes the vocabulary size. Position embeddings $\mathbf{p} = (\mathbf{p}_1,...,\mathbf{p}_n)$ are then added to word embeddings to give a sense of order.  $\mathbf{p}_i \in \mathbb{R}^d$ is defined as:

\begin{equation} \label{eq5}
\begin{split}
p(_{i,2j})&=sin(pos/10000^{2j/d})\\
p(_{i,2j+1})&=cos(pos/10000^{2j/d})
\end{split}
\end{equation}

\noindent where $i$ denotes the position and $j$ denotes the dimension. 

Next component is a stack of $L$ identical layers, each layer is composed of two sub-layers: multi-head self-attention layer and position-wise fully connected feed-forward layer. Multi-head self-attention layer is a special case of MHA layer, its three inputs are all the same $Q=K=V$, and all come from the output of the last layer. 

The output of the encoder has two parts, one is the new word representations, $\mathbf{w}=(\mathbf{w}_{1},...,\mathbf{w}_{n})$, where $\mathbf{w}_{i} \in \mathbb{R}^{d}$; the other is the sentence representation $\mathbf{s} \in \mathbb{R}^{d}$ which is averaged through each time step of $\mathbf{w}$. 

~\\
\noindent \textbf{Decoder} is shown in the right part of Figure 1. We use a three-step strategy to generate action: (1) generate the initial feature of movement at each time step, (2) fine tune these features and (3) finally generate action through the fine-tuned features. 

The strategy to generate the initial feature is shown in the lower part of the decoder. We first concatenate sentence representation $\mathbf{s} \in \mathbb{R}^{d}$ and latent vector $\mathbf{z} \in \mathbb{R}^{d_z}$, where $\mathbf{z}$ is sampled from Gaussian distribution $N(0,1)$. Introducing noise helps to generate diverse actions given one caption, which in turn addresses the one-to-many property of the dataset as described in early section. A fully connected layer is followed to generate a query vector for each time step. We also add position embeddings to all queries. Then a MHA layer and a PFCF layer together generate the initial movement features $\mathbf{h} = \mathbf{h}_{1},...,\mathbf{h}_{m}$, where $\mathbf{h}_{i} \in \mathbb{R}^{d}$. The inputs $K$ and $V$ of the MHA layer are the word representations $\mathbf{w}$ from the encoder and $Q$ is the queries equipped with position embeddings.

The second step, shown in the upper part of the decoder, consists of $L-1$ identical layers to fine tune the initial movement features. Each layer has three sublayers: masked multi-head self-attention layer, MHA layer and PFCF layer.  Masked multi-head self-attention layer is also a special case of MHA layer. Its three inputs $Q=K=V$ and all come from the output of the last layer. At the same time, a mask, whose all lower triangle positions are $1$ and other positions are $-\infty$, is applied to the inner product $QK$ to prevent information flowing from a later time step. In MHA layer, both $K$ and $V$ are word representations $\mathbf{w}$ from encoder, $Q$ is the output of the last sublayer. 

After obtaining the latent representation of the time sequence, the final step is another PFCF layer that generates the action from the fine-tuned movement features.

Residual connection and batch normalisation are employed around all the MHA layers and PFCF layers in the encoder and decoder, except the last PFCF layer in the decoder. Layer normalisation in the original paper is replaced by batch normalisation, we argue that batch normalisation solves the over-smooth problem we find in the experiments if using layer normalisation.

\subsection{Discriminator}
In this work, we also propose three discriminators which are inspired by \cite{DBLP:journals/corr/abs-1804-08264}. Caption-action consistency discriminator is the usual one to judge the overall matching between caption and generated action. Pose discriminator and pose transition discriminator play the role as constraints to improve model performance.

\begin{figure} 
  \includegraphics[scale=0.0165]{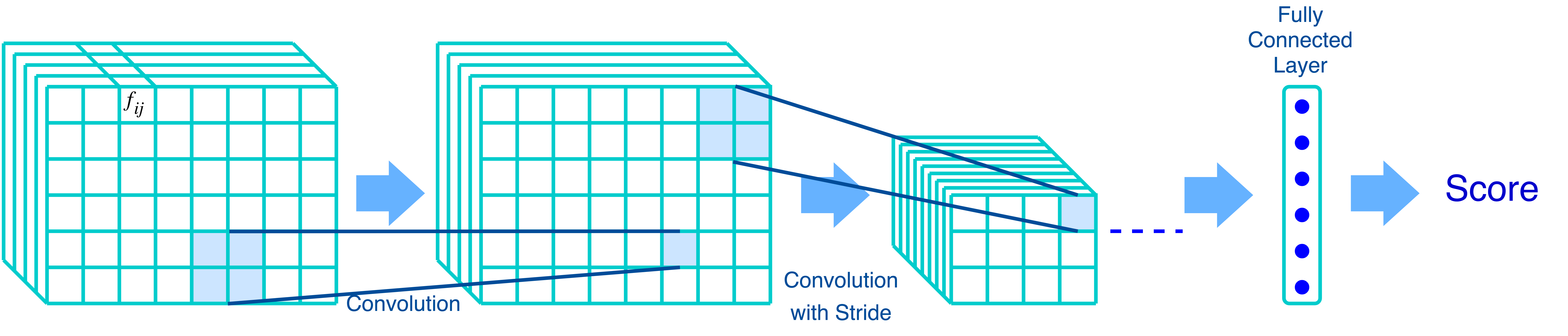}
\caption{Action-Caption Consistency Discriminator}
\end{figure} 

\noindent \textbf{Caption-action consistency discriminator} judges action caption consistency from a holistic perspective, which is shown Figure 3. Similar to \cite{DBLP:journals/corr/WuXZTQLL17}, we first construct a 2D image-like representation $\mathbf{f} \in \mathbb{R}^{n \times m \times (d+d_{f})}$, where $\mathbf{f}_{ij}$ is the concatenation of word embedding at position $i$, and action at position $j$.

\begin{equation} \label{eq6}
\mathbf{f}_{ij} = [\mathbf{e}_{i}; \mathbf{y}_{j}]
\end{equation}

Based on this 2D image-like representation, we employ a set of convolutional layers to capture consistency between caption segments and action segments. Pooling layers are replaced by convolutional layers with stride to prevent sparse gradients flow. Batch normalisation is also employed around each convolutional layer. The detailed layer parameters will be shown in the next section. The extracted caption-action consistency features $\mathbf{f}^{'}$ are flattened and then fed into two fully connected layers to get the probability that caption-action pairs $(\mathbf{x}, \mathbf{y})$ is from the true data distribution:

\begin{equation} \label{eq7}
\begin{split}
D_{1}&(\mathbf{x}, \mathbf{y}) = \\
&Sigmoid(GELU(\mathbf{f}^{'}W_{1}+b_{1})W_{2}+b_{2})
\end{split}
\end{equation}

The first adversarial loss is measured as:

\begin{equation} \label{eq8}
\begin{split}
\mathcal{L}_{1}^{d} &= -[logD_{1}(\mathbf{x}, \mathbf{y}_{real}) + log(1-D_{1}(\mathbf{x}, \mathbf{y}_{fake}))]\\
\mathcal{L}_{1}^{g} &= -logD_{1}(\mathbf{x}, \mathbf{y}_{fake})
\end{split}
\end{equation}

\noindent \textbf{Pose discriminator} aims to discriminate how likely a pose is real given a caption, which is $p(\mathbf{pose}_t | caption)$. As each action has two movement subsequences, each with $m$ time steps, so we have $2m$ poses in total. According to our discussion in Section 2, joints are already ``feature extracted". To better utilise the joints features, we divide the 14 joints into 4 branches to combine these joints features hierarchically. The 4 branches are shown below: 

\begin{itemize}
\item \textit{head, neck, left shoulder, left elbow, left hand}
\item \textit{head, neck, left thigh, left shin, left foot}
\item \textit{head, neck, right shoulder, right elbow, right hand}
\item \textit{head, neck, right thigh, right shin and right foot}
\end{itemize}

\noindent The 3D coordinates of each joint in the branch are concatenated together to construct a vector with dimension $5*3=15$, which is then fed into a fully connected layer to extract the branch feature. Four branch features are extracted with different parameters. We concatenate the four branches feature together with sentence representation to construct the pose-caption feature. The pose-caption feature is fed into two fully connected layers to obtain the probability of a pose being real given a caption. 

The second adversarial loss is measured as:

\begin{equation} \label{eq9}
\begin{split}
\mathcal{L}_{2}^{d} = -\frac{1}{2m} [&\sum_{i=1}^{2m} logD_{2}( \mathbf{x}, \mathbf{pose}_{real}^{i} ) \\
+&\sum_{i=1}^{2m} log(1-D_{2}(\mathbf{x}, \mathbf{pose}_{fake}^{i}))] \\
\mathcal{L}_{2}^{g} = -\frac{1}{2m}&\sum_{i=1}^{2m} logD_{2}(\mathbf{x}, \mathbf{pose}_{fake}^{i}) \\
\end{split}
\end{equation}

\noindent \textbf{Pose transition discriminator} is to discriminate how likely the next pose is real given a caption and a pose at any time step, which is $p(\mathbf{pose}_{t+1} | \mathbf{pose}_t, caption)$. First, we calculate the motion between two adjacent poses in time step, $\mathbf{m}_{t}=\mathbf{pose}_{t+1} - \mathbf{pose}_{t}$. Branch feature extraction method which described in pose discriminator is also applied to each motion branch $\mathbf{m}_{t}^{branch_i} \in \mathbb{R}^{3*5}$ and pose branch $\mathbf{pose}_{t}^{branch_i} \in \mathbb{R}^{3*5}$. Sentence representation, extracted four branch pose features and four branch motion features are concatenated, then be fed into two fully connected layers to obtain the conditional pose transition probability.

The third discriminator loss is measured as:

\begin{equation} \label{eq10}
\begin{split}
\mathcal{L}_{3}^{d} &= -\frac{1}{2(m-1)} [\\
&\sum_{i=1}^{2(m-1)} logD_{3}( \mathbf{x}, \mathbf{pose}_{real}^{i},  \mathbf{m}_{real}^{i}) \\
+&\sum_{i=1}^{2(m-1)} log(1-D_{3}(\mathbf{x}, \mathbf{pose}_{fake}^{i},  \mathbf{m}_{fake}^{i}))] \\
\mathcal{L}_{3}^{g} &= -\frac{1}{2(m-1)}\\
&\sum_{i=1}^{2(m-1)} logD_{3}(\mathbf{x}, \mathbf{pose}_{fake}^{i},  \mathbf{m}_{fake}^{i}) \\
\end{split}
\end{equation}

\subsection{Overall Adversarial Loss}
The overall adversarial loss of our training architecture is summarised as below:

\begin{equation} \label{eq11}
\begin{split}
\mathcal{L}_g = \frac{1}{3}[\mathcal{L}_{1}^{g}+\mathcal{L}_{2}^{g}+\mathcal{L}_{3}^{g}]\\
\mathcal{L}_d = \frac{1}{3}[\mathcal{L}_{1}^{d}+\mathcal{L}_{2}^{d}+\mathcal{L}_{3}^{d}]
\end{split}
\end{equation}

\subsection{Teacher Forcing}
As reported in \cite{NIPS2016_6099, DBLP:journals/corr/WuXZTQLL17, DBLP:journals/corr/YangCWX17}, teacher forcing is an import factor to successfully train a sequence generation model in an adversarial training architecture. Teacher forcing plays the role of a regulariser which significantly improve stability in the training process. In our work, we propose a modified teacher forcing. For a given caption $\mathbf{x}$, the generator G may generate different actions with different latent vector $\mathbf{z}$. We calculate the squared Euclidean distance between generated action $G(\mathbf{x}, \mathbf{z})$ and all real actions corresponding to the given caption. Only the minimum loss value is chosen as the teacher forcing loss:

\begin{equation} \label{eq12}
\begin{split}
\mathcal{L}_{tf} &= min(Dist(G(\mathbf{x}, \mathbf{z}), \mathbf{y}^{[i]})) \\ 
i &\in {1,2,...}
\end{split}
\end{equation}

\noindent The maximum value of $i$ depends on how many real actions corresponding to the given caption $\mathbf{x}$ and \textit{Dist} denotes the squared Euclidean distance.

The training process is shown in the Algorithm 1.

\begin{algorithm}
  \caption{Action Generation From Captions}
  \begin{algorithmic}[1]
  \For{t = 1 to T}
      \State Sample a batch of caption-action pairs \{$\mathbf{x}, \mathbf{y}$\}.
      \State Sample a batch of random noise \{$\mathbf{z}$\}.
      
      \For{d-steps}
      \State Generate fake actions $\mathbf{\hat{y}} = G(\mathbf{x}, \mathbf{z})$.
      \State Calculate D loss via $\mathcal{L}_{d}$.
      \State Update all parameters in Discriminator.
      \EndFor
      
      \For{g-steps}
      \State Calculate Generator loss via $\mathcal{L}_g$.
      \State Update all parameters in Generator.
      \EndFor

      \For{teacher-forcing-steps}
      \State Calculate teacher forcing loss via $\mathcal{L}_{tf}$.
      \State Update all parameters in Generator.
      \EndFor
  \EndFor
  
  \end{algorithmic}
\end{algorithm}

\section{Experiments}
In this section, we elaborate all the experiments details, including data processing, model parameters, experiments and generation performance.

\subsection{Data Processing}
As stated in the section 2, our data set contains caption-action pairs. Each caption is a sequence of words to describe the interactive movement and each action consists of two subsequence, one for each person. 

Each caption is simply tokenise into a sequence of words and the stem of each word is extracted as inputs of the model. The operation of extracting word stem is to limit the size of the vocabulary, we consider there's no difference among words like \textit{walk}, \textit{walks} and \textit{walking}.

Action data is collected with $120$ frames per second, we sample each action uniformly to have $3$ frames per second and limit the maximum length to $26$. For a given action, it consists of two subsequence $\mathbf{y}^{j}=(\mathbf{y}_{1}^{j},...,\mathbf{y}_{m}^{j})$, $\mathbf{y}_{i}^{j} \in \mathbb{R}^{3\times14}$  and $j \in {1,2}$. Firstly, we calculate the centre point $\mathbf{c}_{i}^{i} \in \mathbb{R}^3$ of each pose $\mathbf{y}_{i}^{j}$ in subsequence by averaging the 3D coordinates of \textit{left thigh} and \textit{right thigh}. Then, normalise each pose by subtracting its corresponding centre point. Through this normalisation operation, all same poses at different locations of the 3D space are converted to one same pose at the origin (angle information retained). To keep relative position information between two subsequence, which includes distance and movement direction, we construct a position sequence $\mathbf{d}=(\mathbf{d}_{1},...,\mathbf{d}_{m})$, where $\mathbf{d}_{i} = \mathbf{c}_{i}^{2} - \mathbf{c}_{i}^{1} \in \mathbb{R}^{3}$. The two pose subsequences and position sequence are normalised again separately to have $0$ mean and standard deviation.

Now, we have two normalised subsequence $\mathbf{\bar{y}}^{1}=(\mathbf{\bar{y}}_{1}^{1},...,\mathbf{\bar{y}}_{m}^{1})$, $\mathbf{\bar{y}}^{2}=(\mathbf{\bar{y}}_{1}^{2},...,\mathbf{\bar{y}}_{m}^{2})$ and one position sequence $\mathbf{d}=(\mathbf{d}_{1},...,\mathbf{d}_{m})$.  The final processed action sequence is $\mathbf{t} = (\mathbf{t}_{1},,,.,\mathbf{t}_{m})$, where $\mathbf{t}_{i} \in \mathbb{R}^{14*3*2+3}$ is the concatenation of $[\mathbf{\bar{y}}_{1}^{i}, \mathbf{\bar{y}}_{2}^{i}, \mathbf{\bar{d}}^{i}]$, and will be feed into the model as real actions.

\subsection{Model Parameters and Optimisation}
In experiments, we set word embedding dimension, model dimension and latent vector dimension all to $128$. The number of attention layer $L=4$, number of heads $H=8$ each with dimension $d_{h}=64$. Consider the size of our dataset and vocabulary size, we set our model parameters to a relatively small value compared to Transformers.

All parameters are initialised from Gaussian Distribution $N(0, 0.02)$. We employ vanilla GAN to train our model, learning rate is set to $5 \times 10^{-6}$ as we find a larger learning rate leads to instability very quickly. Adam optimiser is employed with beta1 equals 0.5.

\subsection{Experiments Setup}
Our experiments explore two datasets: one-to-one dataset and one-to-many dataset. For one-to-one dataset, latent vector $\mathbf{z}$ is ignored as each caption has only one unique corresponding action. 
And we also evaluate the impact bring by some model components such as teacher forcing and batch normalisation. 

\subsection{Results}
Figure 3 shows the generated action samples in different experiment setups. Model without teacher forcing can not generate actions with plausible quality even on training data, once again it proves what a significant role the teacher forcing is playing in an adversarial training example. Layer normalisation is not a good choice together with multi-head self-attention layer. We argue that, the essence of self-attention is a weighted combination of projected values, in another word, some kind of smooth effect. However, any two adjacent time step in target sequence has only a very small motion, smooth effect exerting on such sequence leads to frame collapse problem, which means all frames become almost the same. Here, layer normalisation can not solve this problem, but if we replace it with batch normalisation, then problem solves.

\section{Conclusion and Future Work}
In this work, we introduced a task to explore conditional continuous sequence generation together with an action generation dataset. Experiments of our proposed model shows that the boundaries of continuous sequence generation falls behind compared with discrete sequence generation due to the nature of continuous sequence. In future work, a general evaluation framework is in demand instead of human evaluation. And we would like to continue expand our dataset to explore more real situation.

\bibliography{acl2019}
\bibliographystyle{acl_natbib}

\end{document}